# Applying Data Mining and Machine Learning Techniques to Submarine Intelligence Analysis

Ulla Bergsten[1], Johan Schubert[2], and Per Svensson[2]

[1] Department of Assessment Models and Simulation, [2] Department of Information System Technology
Defence Research Establishment, SE-172 90 Stockholm, Sweden
{bergsten, schubert, pers}@sto.foa.se

**Abstract**

We describe how specialized database technology and data analysis methods were applied by the Swedish defense to help deal with the violation of Swedish marine territory by foreign submarine intruders during the Eighties and early Nineties. Among several approaches tried some yielded interesting information, although most of the key questions remain unanswered. We conclude with a survey of belief-function- and genetic-algorithm-based methods which were proposed to support interpretation of intelligence reports and prediction of future submarine positions, respectively.

## Introduction

In 1980, for a period of several weeks the Swedish navy hunted what it later judged to be two foreign submarines operating in the country's inner territorial waters, near Sweden's largest naval base.

This event commenced a more than decade-long period of political uneasiness and increasing military, as well as public, vigilance. The period was characterized by an inflow of final event intelligence reports to the Swedish military headquarters, which during the years of 1986-88 reached a peak of about 1000 per year.

Until the Submarine Commission report (Ubåtsfrågan 1981-1994 1995) was published, very little was known to the public about the size and character of the intelligence material that had been gathered. The work to be described below started in 1986, then considered a top secret activity whose mere existence could not be revealed.

## The report of the 1995 Submarine Commission

In February 1995, the Swedish government formed an independent commission "with the task of assessing and analyzing the underwater violations and indications of these that have existed since the beginning of the 1980's..." (*ibid*).

Most of the collected reports, roughly 80%, are of human observations. Of these, 80% were made by civilians. With regard to these reports, the commission states that "in our opinion, credible observations of foreign submarine activity have been made". The more than 6000 reports were classified by the defense authorities in four quality categories, plus the categories "No submarine activity" and "Not decidable". More than 1500 reports claim a target distance of less than 100 m. Of these, about 400 had been classified as belonging to the categories 1-3, and 40 to category 1.

The Commission report declares that "in our opinion, it is not possible to state the number of credible observations, and, by doing so, to draw a line between these and other observations". In the classification scheme used by the military, the top category "Confirmed activity" was intended to include only such observations that were provably true in a legal sense. As the above citation shows, the relevance of this classification was rejected by the commission.

## The Database and its Toolset

Late in 1986, the naval intelligence analysts had tried a regular C3I system but found it too inflexible for their purpose. By coincidence, our group was able to offer them unique new technology, a system called Cantor (Svensson and Neider 1991, Andersson 1992, Karasalo and Svensson 1986).

Cantor is designed to efficiently manage, analyze, transform, and visualize large sets of data, including spatially and spatio-temporally distributed point observations. It has many properties which make it easier to use for a small group of analysts than a mainstream relational DBMS, such as simple and logical means to define, populate, and display object types and values. It also possesses a more powerful data model and query language (called SAL) than the SQL standard of the late 1980's, allowing, e.g., scalar, tuple, and set-valued objects, parameterized views, and view materialization.

The database input was organized as a collection of measurement tables, containing general observation metadata, such as time, area, position, observation quality, id number etc., observation type data, such as submarine, diver, waves, etc., and detailed observation features, such as size, speed, and heading, shapes, colors, sound character, lighting character, and bottom track characteristics.

## Groping for the Right Questions

The submarine intrusion problem represents a class of data analysis problems where observations form a complex structure, in relation to which it is unknown where and how to find useful information.

On the lowest level of aggregation one faces, e. g., prob-lems of constructing possible paths from uncertain point data, of "counting" the number of targets using only indirect evidence such as time and distance in relation to probable speed, and of finding indirect support for target detection from coincident, more easily observable processes such as radio signals from non-submarine sources. On a more aggregated level, one wants to find spatio-temporal patterns that might be used to predict future behavior.

Below, we will briefly describe techniques for mapping both point data and aggregate spatial information, for visualizing and analyzing statistically the temporal pattern of observations, and for discovering and analyzing clusters in space and time.

One of our first tasks was to show how various observation categories were distributed along the Swedish coastline. Fig 1 depicts the coastline curvilinear distribution of all observations made during a certain time period. The highest peak corresponds to observations in the Stockholm archipelago. In general, the peaks in Fig 1 seem to correspond to areas of naval interest rather than major population centers.

Next, the two-dimensional distribution of observations was mapped as shown in Fig 2. To highlight "hot spots" non-linear pixel coloring was chosen. Stationary acoustical and magnetic sensors are scarce resources which were often installed to guard passages into areas of particular defense interest, commonly giving rise to such hot spots. Since these sensors were more or less on constant alert, they may provide a measure of the temporal distribution of visits to such areas.

A key question when analyzing data from suspected intrusions is whether the observations form a non-random distribution over time (see next section). To begin investigating such questions, the graph of Fig 3 was produced.

Each of these diagrams represent a large family of possible visualizations since the data being displayed can be selected at will, e.g., to show only high-quality civilian sightings of submarine type, or reports from stationary sensor installations.

## Statistical Analysis of the Database

A shallow statistical analysis of the database was performed. It was based on human observations of the categories 1-3 in the period 1986-1991. This set includes roughly 800 reports. The purpose of the statistical analysis was to examine whether the observations occur randomly in time and space. Even if they do not, other explanation factors have to be eliminated before we are able to draw the conclusion that the set of observations arises from foreign submarine intruders.

Cluster formations in the time dimension may arise during summer holidays and weekends because more people are then visiting the archipelago, possibly leading to an increase in the number of observations (whether true or false). Clusters in the spatial dimension may arise in areas with many observers. A mass media effect may also be present, i.e. individuals may show a greater tendency to observe and report phenomena in the sea when mass media have announced an ongoing suspected submarine activity.

To examine if the observations fall randomly in time two kinds of tests were performed. The first test was based on the exact time the observation was made and provided the strongest test results. The second test considers instead the time between successive observations. Jointly, the two tests provide additional information.

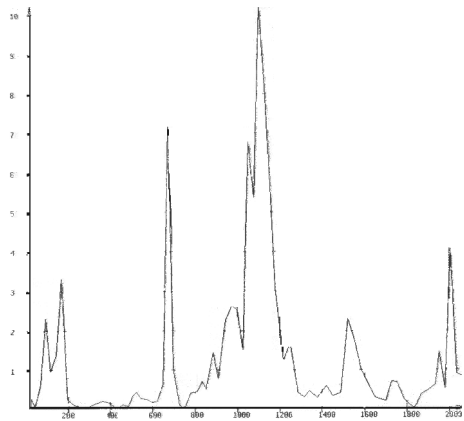

Fig 1 Observation density as a function of arclength along Sweden's coastlline

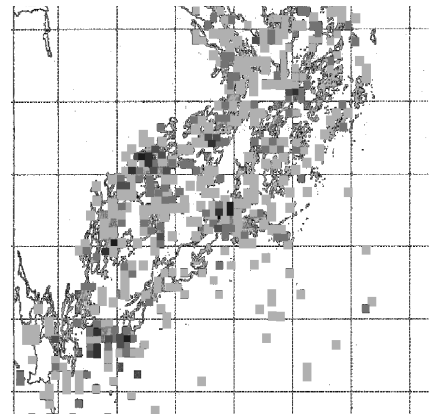

Fig 2 Map of the two-dimensional distribution of of observations within the southern archipelago of Stockholm

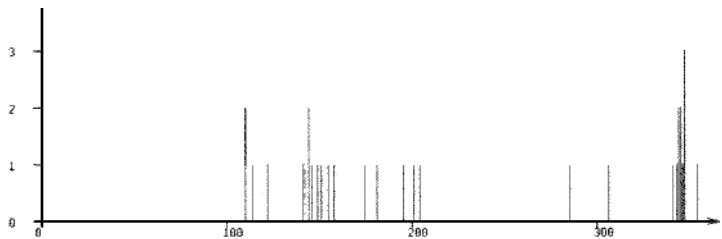

Fig 3 Number and type (originally color coded) of observations per day during one year

For each seasonal period considered, the result stated that the hypothesis of randomness should be rejected.

After having established that the observations do not represent a random distribution, it remains to find out which factors may have caused the non-randomness.

To determine the influence of the number of observers in the archipelago the hypothesis "observations occur with the same frequency workdays and weekends" was tested and consistently accepted.

Another task was to examine, if, when, and where there are clusters in time or space. For this purpose the cluster analysis program CLUSTAN (Wishart 1987) was used. To take into account the fact that several intrusions may be going on simultaneously, and that the same area may be visited more than once during any given time period, the observation set was visualized in two different ways:

1. The observations were clustered with respect to space and the observations within each cluster were shown on a time scale.
2. Similarly, clustering first with respect to time, then to space was studied.

The cluster analyses show that well supported conclusions regarding simultaneous clustering in time and space require a very large and reliable data set. Even if that requirement is not met, however, such studies may provide valuable indications and input to further analysis.

## Using Evidential Analysis to Associate Intelligence Reports

When several similar submarines are operating concurrently, reports never tell which submarine they refer to. Therefore, methods are needed which enable an analyst to separate the intelligence reports into subsets according to which submarine they are referring to (Schubert 1993). Having applied this method, one can then analyze the reports for each submarine separately, e.g., using methods described in (Bergsten and Schubert 1993).

To treat this problem, we use the concept of *conflict* in Dempster-Shafer theory (Shafer 1976) between the propositions of two intelligence reports as a measure of the probability that the two reports are referring to different submarines.

In Fig 4 these subsets are denoted by $\chi_i$ and the conflict when all pieces of evidence in $\chi_i$ are combined by Dempster's rule is denoted by $c_i$. When the number of subsets is uncertain there will also be a "domain conflict" $c_0$ which is a conflict between the current hypothesis about the number of subsets and our prior belief.

The cause of the conflict can be non-firing sensors placed between the positions of the two reports, the required velocity to travel between the positions of the two reports at their respective times in relation to the assumed velocity of the submarines, etc.

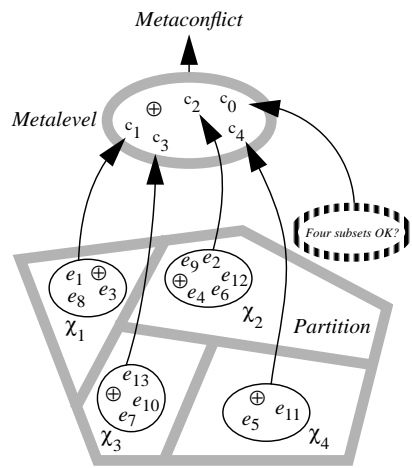

Fig 4 The conflict in each subset of the partition becomes a piece of evidence at the metalevel

We use the minimizing of a criterion function of overall conflict (the *metaconflict* function) as the method of partitioning the evidence into subsets representing the events.

The method of finding the best partitioning is based on an iterative minimization of the metaconflict function. In each step the consequence of transferring a piece of evidence from one subset to another is investigated. After this, each subset of intelligence reports refers to a different target and the reasoning can take place with each target treated separately.

We may also specify each piece of nonspecific evidence by observing changes in cluster and domain conflicts when moving a piece of evidence from one subset to another (Schubert 1996). Without this extension the most plausible subset would take this piece of evidence as certainly belonging to the subset.

Finally, we established a posterior probability distribution regarding the number of subsets (Schubert 1995).

## Making Tactical Predictions from Learned Patterns

We developed a machine-learning system for making short-term predictions, based on methods which recognize an incoming sequence of intelligence reports as belonging to a certain category of sequences. Having found such a category we obtain probabilities for different future developments given the current situation.

By use of a genetic algorithm, our system learns the categories of sequences of (simulated) intelligence reports. When we receive a new scenario it is analyzed using the learned categories. If the system finds a category of sequences with a beginning similar to the current sequence, the remainder of the historical sequences are used to give a prediction about the future.

We show in Fig 5 a sequence of three intelligence reports. The latest report from area E5 is placed to the right on the time scale at T0, and the two earlier reports from area I4 and F4 are placed in their respective time intervals, T5 and T3.

In learning we will now try to predict the next event. Suppose it will happen in area E7 two time intervals into the future, Fig 6. Our aim is to find a rule that predicts this event based on the earlier information. Such a rule can be highly specific in both precedent and prediction:

**If** [I4 & T5] [F4 & T3] [E5 & T0] **then** [E7 & T-2],
or it may throw a much wider net, e.g.:
**If** [HIJ345 & T456] [EF34 & T1234] [DEF456 & T0] **then** [CDEF67 & T-1-2], Fig 7.

Both alternatives and all other combinations with a certain specificity in the precedent, or parts of it, and another specificity in the prediction are possible and automatically tested during the learning phase. The disadvantage of a specific prediction is that the prediction rule tends to become a special case and may also get a low probability. A less specific rule has a higher probability but is not as useful in the individual case. The learning mechanism uses a scoring method that takes this into account and finds a suitable balance.

A statistical analysis based on a simulation of the method showed that the probability of a correct prediction was at best 54%, with an accuracy in predicted position of 5 kilometers and in predicted time of 48 minutes. Prediction rules with a probability and an accuracy such as these should be very useful if they can be approached in practice.

## Conclusions

The submarine intelligence database contains a huge amount of information of varying credibility. Even if a large part of the data is of low credibility, this part complements the picture obtained from the more reliable data.

The specialized knowledge and software technology needed for state-of-the-art data mining and analysis needs to be mastered by the analyst group itself rather than by their consultants, in particular when dealing with top-secret information. To be able to exploit these complex techniques fully, analysts need to be fully in charge of their work and its tools.

Much of the power of data analysis lies in the opportunity to explore, more or less immediately, any promising idea that is generated by the mind of the analyst. In the absence of such dedicated, full-time access, the work reported here has only scratched the surface of a veritable mountain of information.

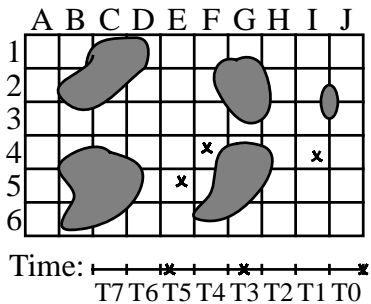

Fig 5 Representation

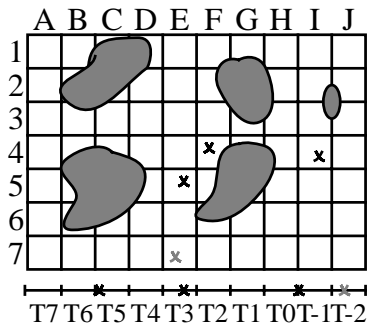

Fig 6 Find a rule to predict the next report

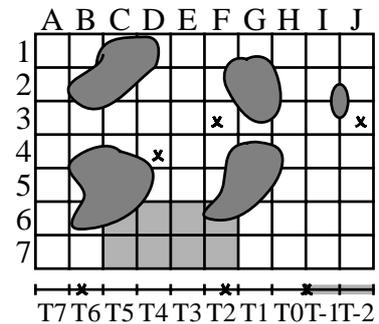

Fig 7 A prediction rule makes a prediction about an area and a time interval